\pdfoutput=1

\documentclass[11pt]{article}

\usepackage[final]{acl}

\usepackage{times}
\usepackage{latexsym}

\usepackage[T1]{fontenc}

\usepackage[utf8]{inputenc}

\usepackage{microtype}

\usepackage{inconsolata}

\usepackage{tcolorbox}
\usepackage{graphicx}
\usepackage{booktabs} 
\usepackage{multirow}
\usepackage{float}
\usepackage{xspace}
\usepackage{subfig}

\definecolor{amaranth}{rgb}{0.9, 0.17, 0.31}
\definecolor{awesome}{rgb}{1.0, 0.13, 0.32}
\definecolor{bittersweet}{rgb}{1.0, 0.44, 0.37}

\newcommand{\ourwork}{\textsc{PatentWriter}\xspace}
\title{\ourwork: A Benchmarking Study for Patent Drafting with LLMs}

\author{Homaira Huda Shomee \\ University of Illinois Chicago  \\ hshome2@uic.edu
        \And  Suman Kalyan Maity \\\parbox{4.2cm}{\centering Missouri University of Science and Technology} \\smaity@mst.edu \And
        Sourav Medya \\University of Illinois Chicago \\medya@uic.edu }

\begin{document}
\maketitle
\begin{abstract}
Large language models (LLMs) have emerged as transformative approaches in several important fields. This paper aims for a paradigm shift for patent writing by leveraging LLMs to overcome the tedious patent-filing process. In this work, we present \ourwork, the first unified benchmarking framework for evaluating LLMs in patent abstract generation. Given the first claim of a patent, we evaluate six leading LLMs—including GPT-4 and LLaMA-3—under a consistent setup spanning zero-shot, few-shot, and chain-of-thought prompting strategies to generate the abstract of the patent. Our benchmark \ourwork goes beyond surface-level evaluation: we systematically assess the output quality using a comprehensive suite of metrics—standard NLP measures (e.g., BLEU, ROUGE, BERTScore), robustness under \textit{three} types of input perturbations, and applicability in \textit{two} downstream patent classification and retrieval tasks. We also conduct stylistic analysis to assess length, readability, and tone. Experimental results show that modern LLMs can generate high-fidelity and stylistically appropriate patent abstracts, often surpassing domain-specific baselines. Our code and dataset are open-sourced to support reproducibility and future research.
\end{abstract}



%

\section{Introduction \& Related Work}
Patents provide a legal framework to protect intellectual property and play an essential role in fostering innovation. For technological advancement, they not only recognize inventors' creativity but also incentivize further innovation by granting them the sole authority to profit from their creations. At the heart of the patent process lies the task of patent writing which has been characterized by its meticulous and time-consuming nature \cite{roberts2007modern,mehta2017inventions,trappey2020intelligent}. This often requires extensive legal knowledge, technical expertise, and linguistic precision \cite{risch2021patentmatch}. It involves crafting detailed descriptions of inventions, drafting comprehensive claims, and ensuring compliance with intricate legal standards—all of which can present formidable challenges for inventors and patent attorneys alike. However, the emergence of Large Language Models (LLMs) give us an opportunity to ease some of these burdens and streamline the patent-drafting process.

LLMs represent a significant milestone in NLP research as they offer advanced capabilities in understanding and generating human-like text. They have demonstrated versatility and effectiveness in generating coherent and contextually relevant text across various domains. For instance, in healthcare, LLMs have been used for generating biomedical text~\cite{peng2023study}, such as summarizing medical literature~\cite{beltagy2019scibert}, generating clinical notes, and composing drug labels~\cite{goel2023llms}. They have also shown promise in diagnostics, clinical decision support, drug discovery, and patient communication~\cite{liu2025application}. In finance and economics, LLMs have been deployed for generating financial reports and economic forecasts~\cite{liu2021finbert,yang2023fingpt}, as well as for supporting financial decision-making tasks such as trading, portfolio management, and risk assessment~\cite{fincon}. In social media, LLMs have been used for hate speech detection~\cite{guo2024investigation} and misinformation mitigation~\cite{chen2024combating}. LLMs also offer significant opportunities in education, particularly for students as aids in research and academic writing~\cite{kasneci2023chatgpt}, interactive study guides with activities such as generating practice questions and delivering instant feedback~\cite{tate2023educational}. 

\textbf{Patent Domain. }With a huge promise, LLMs have also started to gain attention in the patent domain, especially in automating some aspects of the patent drafting process \cite{krestel2021survey,lee2020controlling,lee2020patent}. An early study in this domain is the PatentTransformer\cite{lee2020patent2}, which employs a GPT-2-based architecture trained on patent data to generate patent segments. \cite{christofidellis2022pgt} introduce the Patent Generative Transformer (PGT), a transformer-based multitask language model designed to streamline the patent generation process through tasks such as part-of-patent generation. 
PatentGPT \cite{ren2024patentgpt} introduces cost-efficient large language models trained on 240B IP-related tokens to support tasks like patent drafting and translation. It uses a two-stage pretraining approach and aligns the models using supervised fine-tuning (SFT) and reinforcement learning from human feedback (RLHF). \cite{jiang-etal-2025-large} evaluate various LLMs for patent claim generation and find that generating claims from detailed patent descriptions yields better results than using abstracts. Interestingly, general-purpose models like GPT-4 outperform domain-specific patent models. AutoPatent \cite{wang2024autopatent} introduces a multi-agent framework that uses planning, writing, and reviewing agents to generate complete high-quality patents from inventor drafts.

\textbf{Our Contributions.} 
While several recent works have explored using LLMs for patent generation, it is difficult to compare the findings because of the wide variations in the datasets, tasks, and evaluation techniques used. In contrast, we provide a unified and controlled benchmarking framework: we evaluate multiple state-of-the-art LLMs using the same dataset, identical task formulation, and standardized prompts. We also conduct a comprehensive evaluation that includes NLP-based similarity metrics, domain-specific tasks (classification, retrieval), robustness under input perturbation, and stylistic quality assessment. \textit{To the best of our knowledge, this is the first benchmarking study in the domain of patent generation using a unified framework.}
Our main contributions are as follows: 
\begin{itemize}
    \item \textbf{LLM variants.} We benchmark the capabilities of six leading LLMs variants in drafting patent documents automatically under a unified setup. In particular, we generate the abstract using the first claim (Secs. \ref{sec:gen_abstract}).
    \item \textbf{Robustness. }We explore multiple prompting strategies--- such as zero-shot, few-shot, and chain-of-thought---and evaluate model robustness under realistic perturbations such as typos, contextual substitutions, and word swaps. This provided insights into the resilience of model outputs in noisy real-world scenarios.
    \item \textbf{Comprehensive evaluation measures: NLP \& domain-based. }We build a comprehensive evaluation measures beyond standard NLP metrics (e.g., BLEU, ROUGE, BERTScore) and assess the practical domain utility of generated texts using two downstream tasks: patent classification and patent retrieval (Secs. \ref{domain:classification} \& \ref{domain:ret}) .
    \item \textbf{Qualitative analyses. }We conduct a qualitative and stylistic analysis of LLM-generated patent abstracts such as length, readability, and passive voice usage (Secs. \ref{sec:qual}). 
\end{itemize}


\begin{figure}[t]
\centering
\includegraphics[width=\columnwidth]{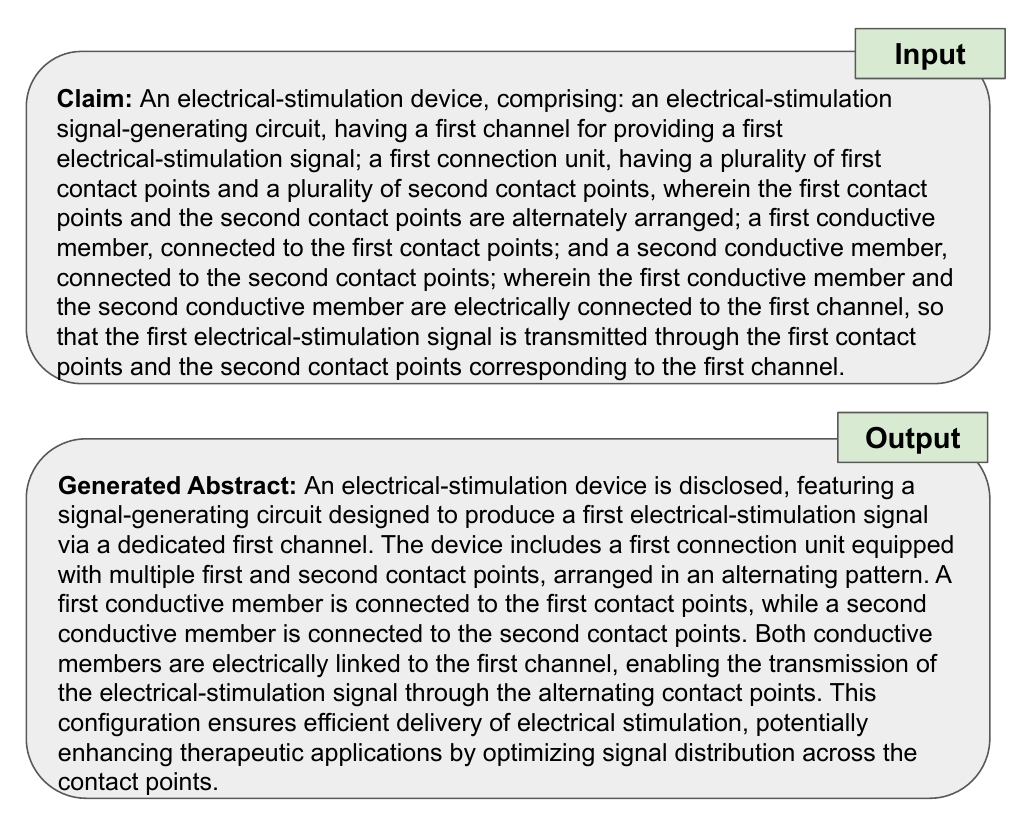}
\caption {\label{fig:example}Example of  Input (Claim1), and Generated Output (Abstract) by GPT4o.}
\end{figure}

\section{Problem of Patent Writing}
\label{sec:problem_def}
A patent typically contains a large volume of content and requires significant human efforts~\cite{roberts2007modern}. Automating the patent drafting process can significantly reduce the time, effort, and legal requirements involved. It can also save costs by reducing the amount of time required from patent attorneys. 
Patent drafting involves using (e.g., prompting) an LLM to generate specific sections of a patent, such as the abstract, independent claims, etc. The generation process aims to accurately describe the invention where patent documents require the use of precise and technical language \cite{risch2021patentmatch}. \\
\textbf{Abstract and the first claim. }Patent claims and abstracts are key components of the patent application. The first claim is arguably the most important part of a patent. It defines the scope of protection sought for the invention. Patent claims outline the specific features and characteristics that distinguish the invention from existing technologies ~\cite{mehta2017inventions}. As such, the first claim serves as a concise summary of the invention's key elements and establishes the boundaries of the patent's legal protection. It is essential for defining the invention's novelty and inventiveness, and it significantly influences the patent's enforceability against infringement and commercial value. 

On the other hand, the abstract provides a brief overview of the invention described in the patent application. It summarizes the technical field, the problem addressed by the invention, its solution, and its advantages. It is typically used by patent examiners, potential licensees, investors, and competitors to quickly grasp the essence of the invention without delving into the detailed description~\cite{handbookIP}. Moreover, the abstract is often published alongside the patent application, making it one of the first things that individuals obtain while searching patent databases.

\textit{In this benchmark, our main objective is generating abstract given the first claim.} The framework can be extended for other inputs and outputs. Figure \ref{fig:example} shows an example where the input is a patent claim, and the output is the corresponding abstract generated by GPT-4o.

\section{Our Benchmarking Framework: \ourwork}
\begin{figure*}[t]
\centering
\includegraphics[width=0.85\textwidth]{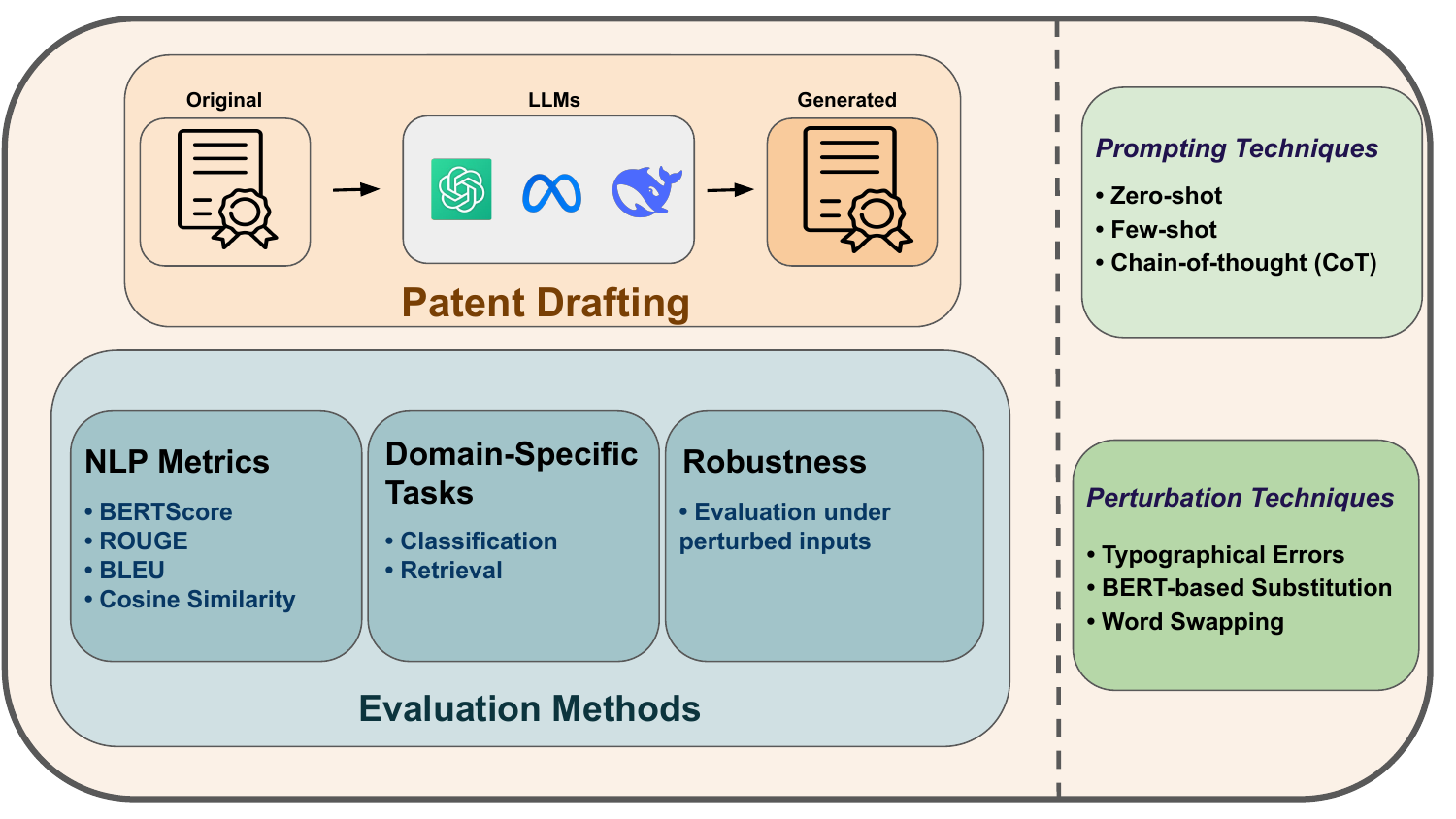}
\vspace{-4mm}
\caption {\label{fig:framework}Overview of \ourwork for assessing abstract generation from patent claims using large language models (LLMs). The left block shows the input–output setting where patent claims are used to generate abstracts through 6 LLMs variants. The generated outputs are evaluated using three key dimensions: (1) \textbf{4} NLP metrics such as BERTScore, ROUGE, BLEU, and Cosine Similarity to measure surface-level and semantic similarity; (2) \textbf{2} domain-specific task performance like classification and retrieval accuracy; and (3) Robustness analysis, which measures the consistency of model outputs under \textbf{3} input perturbations. The right block represents variations in prompting strategies—zero-shot, few-shot, and chain-of-thought (CoT)—as well as perturbation techniques applied to the input, including synthetic typing errors, BERT-based contextual replacements, and word swaps. }
\vspace{-3mm}
\end{figure*}

We propose a comprehensive patent benchmarking framework \ourwork to assess the quality of LLM-based patent text generation. Figure \ref{fig:framework} shows the detail of our benchmarking framework. We outline the components of \ourwork below. 
\subsection{Benchmark Dataset}
The dataset is derived from the PatentsView \footnote{https://patentsview.org/download/data-download-tables} and consists of U.S. patents granted in 2022. It includes claim-abstract pairs drawn from 21 CPC subclasses spanning A61 (medical), G06 (computing), and H04 (telecommunications). To ensure balanced coverage, we sample approximately 1,000 instances from each subclass. Each data point contains the patent ID, title, abstract, and corresponding CPC label. Additional details can be found in Appendix \ref{app:data}.
\subsection{Generation by Large Language Models}

\label{approach:genLLM}
Large language models (LLMs) are effective AI assistants that can handle complex reasoning tasks that require expert knowledge in various fields \cite{yang2023fingpt, peng2023study}.  We evaluate the capabilities of multiple LLM backends in different variants such as GPT (including 3.5, 4o, and 4.1), Llama (versions 2 and 3), and DeepSeek models as the generative model to write abstracts from the first claim of the same patent. 
 \\
\subsection{Different Prompting Techniques}
Prompting serves as a fundamental and extensively adopted paradigm for directing the behavior of large language models (LLMs) \cite{brown2020language, liu2023prompting}. Therefore, to systematically assess its impact on generation quality, we evaluate multiple prompting strategies. They are as follows:

\begin{itemize}
\item \textbf{Zero-shot prompt}: This provide the model with only a task description, without any examples. This is the simplest form of prompting and test the generalization ability of the model based on its pretraining knowledge.
\item \textbf{Few-shot prompt}: In this prompt, we include a small number of input-output examples to condition the model on the desired output format and content. This method utilize in-context learning to improve coherence, structure, and adherence to domain-specific language. For instance, we provide three claim-abstract pairs followed by a new claim, prompting the model to complete the corresponding abstract. We choose three examples with varying lengths and performance  levels (high, medium, and low NLP scores).  

\item \textbf{Chain-of-thought (CoT) prompts}: These prompts explicitly instruct the model to reason step-by-step before producing the final output. This helps the model handle tasks that need reasoning or multiple steps.
\end{itemize}
Examples on different prompting techniques are shown in Appendix \ref{app:prompting}.

\subsection{Different Perturbation Techniques}
To evaluate the robustness  of LLMs, we introduce a diverse set of input perturbation techniques. These methods simulate realistic variations or noise in the input (e.g., claims) to test whether the models produce consistent outputs under minor disturbances. We apply both character-level and word-level perturbation using the nlpaug Python library \cite{ma2019nlpaug}. The first perturbation introduces typographical errors via simulated keyboard typos that captures the kinds of accidental errors common in human drafting.
Next, we apply a BERT-based contextual substitution model, which replaces words with contextually appropriate alternatives to push LLMs to handle subtle shifts in language. Another method randomly swaps adjacent words in the input to test the model's sensitivity to mild syntactic disorder. Examples of different perturbation techniques are provided in Appendix  \ref{app:perturbation}.

\subsection{NLP-based Evaluation Metrics}
\label{sec:eval_nlp}
Traditionally, for the evaluation of generated texts, NLP-based measures have been used in the literature. 
The purpose of these metrics is to quantitatively measure different aspects of the quality of the generated text, such as coherence and relevance. The metrics are as follows. (1) \textbf{BERTScore}: BERTScore ~\cite{zhang2019bertscore} evaluates the semantic similarity between the generated text and reference (original) texts using the contextual embeddings. In our framework, this is the major evaluation measure as we also aim for preserving the context accuracy, coherence, preciseness from the original text. In the literature, other NLP-based measures have been used for evaluation. However, they are weaker in the sense that they capture the similarity only via similar patterns in the text. \textbf{ (2) ROUGE-L}: ROUGE-L \cite{lin2004rouge} assesses the longest common subsequence (LCS) between the generated and the reference text. It gauges semantic coherence by computing  \textit{precision, recall, and F1-scores} based on this sequence. \textbf{(3) BLEU}: It measures the overlap of n-grams between the reference and generated text \cite{papineni2002bleu}. Although it is designed for measuring the quality of machine translations, it has since been used in other NLP tasks where generated text needs to be evaluated against a reference or human-generated text since it correlates reasonably well with human judgment. \textbf{(4) Cosine similarity} measures the cosine of the angle between two non-zero vectors in a multi-dimensional space \cite{cosine}.

\subsection{Evaluation on Patent-related Tasks}
\label{sec:eval_patent}
In addition to assessing generated patent documents from an NLP standpoint, we evaluate the usefulness of them in patent-related tasks.

\textbf{Patent Classification.} Patent classification is an important and time-consuming task in the patent life cycle \cite{krestel2021survey}. This task involves a multi-class classification (e.g., CPC) for patents where the classification scheme is hierarchical and a patent can get multiple labels in general. We simplify the problem by considering only the single-label classification setting. We classify the generated patent abstracts into subclasses in a particular class. We consider three classes (Table \ref{tab:class_code_name}) and classify the patents in each class separately. The goal is to compare the classification accuracy of the model between the original and generated abstracts as inputs, rather than to enhance the overall accuracy of patent classification. Our objective is to identify any disparities in classification performance between these two sets of abstracts and thus, evaluate the usefulness of the generated abstract. 

\begin{table}[ht]
\centering

\resizebox{.95\columnwidth}{!}{%
\begin{tabular}{cc}
\toprule
CPC Codes & Categories  \\ \midrule
A61 &Medical or Veterinary Science; Hygiene\\
G06 & Computing; Calculating or Counting \\
H04 & Electric Communication Technique \\
\bottomrule
\vspace{-3mm}
\end{tabular}%
}
\caption{Three major classes used in patent generation. These classes have sub-classes and the details are shown in Appendix (Table \ref{tab:data_class_subclass_full}).}
\label{tab:class_code_name}
\end{table}

\noindent
\textbf{Patent Retrieval.} The patent Retrieval (PR) task focuses on effectively retrieving relevant patent documents given a specific search query \cite{shalaby2019patent}. To evaluate the usefulness of LLM-generated patent abstracts, we design a retrieval-based similarity experiment comparing human-written original abstracts with their generated counterparts. Our  hypothesis is that well-generated abstracts should retrieve a similar set of patents as the original abstract. 
\subsection{Qualitative Measures}
\label{sec:eval_qual}
In addition to quantitative evaluation metrics, we assess the linguistic and stylistic quality of LLM-generated patent abstracts using three qualitative measures. First, we compute the abstract length (in tokens) to check verbosity. Second, we use the  readability score to evaluate the linguistic complexity of the generated text. Third, we measure the percentage of passive voice usage, a common stylistic feature in patent writing. These qualitative metrics provide complementary insights beyond quantitative similarity.

\section{Experimental Results}
We demonstrate the followings: (i) the quality of the generated \textit{abstracts} by several LLMs, (ii) robustness of the generation, (iii) the usefulness of the generated texts for the patent domain, and (iv) qualitative analyses of the generated patent abstracts. 
We also include additional analyses in the appendix. For instance, we explore how abstract length correlates with claim length. It shows that LLM-generated abstracts tend to mirror input verbosity more closely. Additionally, we compare word usage patterns, where generated abstracts exhibit more repetitive and templated phrasing compared to original abstracts. The code is available here: \url{https://anonymous.4open.science/r/pwriter-A95C}


\subsection{Drafting Patent Abstracts}
\label{sec:gen_abstract}

\textit{We generate abstracts of patents by using the first claim of the corresponding patent as an input}.  
Subsequently, we evaluate the quality of the generated abstract while focusing on the similarity between the generated abstracts and the original ones. A high similarity would suggest that LLMs are adept at generating abstracts with greater accuracy, and will potentially lead to significant cost and resource savings. We compare the original and generated abstracts based on the NLP-based measures. The measures,  BERTScore, cosine similarities, ROUGE, and BLEU are shown in Table \ref{tab:NLP_Eval} for the sub-class A61, G06 and H04 and an example (Fig. \ref{fig:example2}) in the Appendix. Note that BERT-based measures are based on semantic similarity between the generated text and the original text using the context-based representations and thus, they are more powerful measures in capturing the similarity between two texts; whereas, other measures are not based on context. For instance, BLEU measures only the overlap of n-grams between the generated and the original text. From these tables, the BERT-based metrics are constantly high (higher is better) across all subclasses. In particular, the BERT score is higher than 0.85 in all cases, goes up to 0.89.  These results indicate a strong performance of LLMs in generating similar abstract as the original one. 

As Llama 3 and GPT-4o produce similar outputs and are efficient among all models in Table \ref{tab:NLP_Eval}), we demonstrate the capabilities of LLM for other tasks in the next experiments using these two models.\\
\textbf{Inference Time.} 
We observed substantial differences in generation time across models. More resource-intensive models, such as LLaMA 3 (8B) and DeepSeek-R1-Distill-Qwen-1.5B, required significantly more time and compute compared to more efficient models like GPT-4o mini and GPT-4.1. A detailed breakdown of inference times and hardware settings is provided in Appendix~\ref{app:time}.

\begin{table}[t]
\centering

\resizebox{.99\columnwidth}{!}{%
\begin{tabular}{@{}lcccccccc@{}}
\toprule
\textbf{Model}& \textbf{CPC} & \textbf{BERT} &  \textbf{Cos} & \textbf{RO} &\textbf{BL}\\ \midrule
Llama 2& A61  &0.87 &0.52 &0.36  &0.12 \\
& G06  &0.89  &0.65  & 0.44 &0.18 \\
 & H04  & 0.89 &0.66  & 0.45 & 0.19\\
\bottomrule
Llama 3 & A61  &0.87 &0.50 & 0.34 & 0.10\\
& G06 &0.88  &0.61  &  0.40&0.14 \\
 & H04  &0.88 &0.62  & 0.41 &0.16 \\
\bottomrule

DeepSeek & A61  &0.85 &0.41 & 0.26 & 0.04\\
& G06  & 0.86 & 0.47 & 0.30 &0.05 \\
 & H04  & 0.87 & 0.50 &0.32  &0.07 \\
\bottomrule
GPT-3.5& A61  & 0.87&0.48 & 0.34 & 0.09\\
& G06  & 0.88 &0.57  & 0.43 & 0.11\\
 & H04  & 0.88 &0.60  & 0.37 &0.10 \\
\bottomrule
GPT-4o & A61  &0.87 &0.49 &0.30  & 0.07\\
& G06  & 0.88 & 0.58 & 0.36 & 0.09\\
 & H04 & 0.88 &0.60  & 0.37 &0.10 \\
\bottomrule
GPT-4.1 & A61  & 0.86& 0.47& 0.30 & 0.06\\
& G06  & 0.87 & 0.55 & 0.34 & 0.07\\
 & H04  & 0.88 &0.57  & 0.35 &0.08 \\

\bottomrule
\end{tabular}%
}
\caption{Evaluation of the generated abstracts under basic prompting using standard NLP-based metrics---BERTScore (BERT), Cosine Similarity (Cos)
, ROUGE (RO), and BLEU (BL)---across three CPC subclasses in the A61, G06 and H04. The  models include Llama 2, Llama 3, DeepSeek, GPT-3.5, GPT-4o, and GPT-4.1. BERTScore remains consistently high across all models and subclasses, indicating strong semantic similarity to the original abstracts.  Llama 3 and GPT-4o  shows competitive performance. Llama 2 shows strong performance across all metrics, especially in Cosine and BLEU and DeepSeek falls short on most metrics.}
\label{tab:NLP_Eval}
 \vspace{-2mm}
\end{table}

\subsection{Impact of Different Prompting Strategies}

 We assess how different prompting strategies and input perturbations influence the quality of LLM-generated patent abstracts on a subset of A61 subclass. We experiment with three prompting methods—zero-shot, few-shot, and chain-of-thought (CoT)—using GPT-4o.  Table \ref{tab:NLP_Eval_2} shows that all three prompting strategies achieve identical BERT and Cosine similarity scores. However, CoT prompting yields higher ROUGE and BLEU scores. This indicates that while all prompting methods effectively capture core content, CoT prompt follows more  of the target style.
 \begin{table}[h]
\centering
\resizebox{.99\columnwidth}{!}{%
\begin{tabular}{@{}lcccccccc@{}}
\toprule
Prompt& Model & BERT &  Cos & RO &BL\\ \midrule
Zero-shot & GPT4o  & 0.87 & 0.48 & 0.30 &0.06 \\
Few-shot & GPT4o  & 0.87 & 0.48 & 0.32 &0.08 \\
CoT & GPT4o  &0.87  &0.49  &0.34  &0.10 \\
\bottomrule
\end{tabular}%
}
\caption{Evaluation of the generated abstracts by the NLP-based measures for the sub-classes in the \textbf{A61 (medical)} class for different prompt techniques. The model used here is GPT-4o.}
\label{tab:NLP_Eval_2}
 \vspace{-2mm}
\end{table}

\subsection{Impact of Input Perturbation}
To evaluate robustness, we introduce three types of perturbations to the input claims: typographical errors, BERT-based contextual word substitutions, and word order swaps. Despite these perturbations, both GPT-4o maintain relatively stable performance, with only modest drops in BLEU and ROUGE scores. This suggests that strong LLMs are not only effective under clean inputs but also resilient to noisy or imperfect user inputs.  Table \ref{tab:NLP_Eval_perturb} shows the performance of GPT-4o under various input perturbation settings. We see that all the measures produce similar results except for slight drop in ROUGE. It shows that these perturbations do not affect the generation process.

\begin{table}[H]
\centering
\resizebox{.99\columnwidth}{!}{%
\begin{tabular}{@{}lcccccccc@{}}
\toprule
Perturbation& Model & BERT &  Cos & RO &BL\\ \midrule

Without pert. & GPT4o  & 0.87 & 0.48 & 0.30 &0.06 \\
Typo & GPT-4o  & 0.86 & 0.47 & 0.28 & 0.06\\
\bottomrule
Bert context. & GPT-4o  & 0.86 & 0.46 & 0.26 &0.05 \\
\bottomrule
Swaps & GPT-4o  & 0.86 & 0.47 & 0.28 & 0.06\\
\bottomrule
\end{tabular}%
}
\caption{Evaluation of the generated abstracts by the NLP-based measures for the sub-classes in the \textbf{A61 (medical)} class for different perturbation techniques. The models used here are Llama 3 and GPT-4o.}
\label{tab:NLP_Eval_perturb}
\end{table}

\subsection{Domain-based Evaluation I: Patent Classification}
\label{domain:classification}
After demonstrating the capability of the LLM in generating high-quality abstracts (Sec. \ref{sec:gen_abstract}), here, our goal is to show \textit{the generated abstracts are indeed useful for domain-related tasks such as patent classification}. The task involves a multi-label classification for patents in a particular subclass. For instance, the patents in class A61 (medical) will be classified into 8 subclasses. We similarly processed H04 and G06 sets, across their respective 6 and 7 subclasses. We fine-tune a transformer-based classifier using these subclass labels as targets and evaluate the model on both human-written and LLM-generated abstracts. 
For a detailed experimental set-up please refer to Appendix \ref{app:claasification_setup}.

Table \ref{tab:classification} shows the results. We observe that GPT-4o consistently outperforms both the original and Llama 3 generated abstracts across most CPC subclasses in terms of precision, recall, F1, and accuracy. In particular, GPT-4o achieves the highest scores in the H04 subclass with an F1 and accuracy of 0.60 and 0.59, respectively. While Llama 3 performs competitively in A61 and H04, its performance slightly drops in G06. Overall, the results suggest that GPT-4o generated abstracts are preserve class specific information better than other LLMs.

\begin{table}[H]
\centering
\small
\resizebox{0.99\columnwidth}{!}{%
\begin{tabular}{llcccc}
\toprule
\textbf{CPC} & \textbf{Type} & \textbf{P} & \textbf{R} & \textbf{F1} & \textbf{Acc} \\
\midrule
\multirow{3}{*}{A61} 
    & Original    & 0.54 & 0.57 & 0.55 & 0.56 \\
    & GPT-4o      &0.58 & 0.60 & 0.57 &0.59  \\
    & Llama 3     & 0.56   & 0.58   & 0.56  & 0.57   \\
\midrule
\multirow{3}{*}{G06} 
    & Original       & 0.54 & 0.53 & 0.53 & 0.53 \\
    & GPT-4o       & 0.56 & 0.56 & 0.56 & 0.55 \\
    & Llama 3    & 0.54   & 0.54   & 0.54   & 0.53   \\
\midrule
\multirow{3}{*}{H04} 
    & Original      & 0.58 & 0.60 & 0.58 & 0.57 \\
    & GPT-4o       & 0.60 & 0.62 & 0.60 & 0.59 \\
    & Llama 3     & 0.59   & 0.60   & 0.59   & 0.58   \\
\bottomrule
\end{tabular}
}
\caption{Classification results on original and LLM generated abstracts across CPC subclasses. GPT-4o and Llama 3 rows show performance using generated abstracts and original abstracts serve as reference. Generated abstracts consistently shows better performance which validates the usefulness of the generated texts.}
\label{tab:classification}
\end{table}

\subsection{Domain-based Evaluation II: Patent Retrieval}
\label{domain:ret}
In this experiment, we aim to validate the generated abstract through another domain-related measure. Here, the domain-related task is patent retrieval (PR). PR plays a crucial role in identifying new patents related to new inventions. It involves efficient retrieval of relevant patent documents for prior art search. Rather than evaluating retrieval performance on some criteria (e.g., class labels), our goal is to assess whether the retrieval behavior of the generated abstract mimics that of the human-written original abstracts. Specifically, we test whether both abstract types retrieve a similar set of patents when used as queries. To that end, we use a Sentence-BERT model to embed each abstract into a dense vector, and compute cosine similarity with all other abstracts in the dataset. Patents are then ranked based on these similarity scores. We then compare the retrieval results of the original and generated versions using overlap@k (for k=5, 10, 25), which quantifies the intersection between their top-k retrieved sets, and Spearman rank correlation, which measures global rank agreement. To measure performance beyond chance, we introduce a randomized baseline where the generated abstracts are shuffled across the dataset before retrieval. For a detailed
experimental set-up please refer to Appendix \ref{app:retrieval_setup}. This setup allows us to evaluate semantic alignment, under the hypothesis that a well-formed generated abstract should retrieve closely related patents—just as the original would.

Table \ref{tab:retrieval} shows the results. GPT-4o consistently outperforms the random baseline and shows slightly higher retrieval similarity than Llama 3 across all CPC subclasses. Notably, GPT-4o achieves the highest Spearman correlation (0.67) and top-k overlaps in the A61 subclass. Llama 3 performs comparably, especially in G06 and H04, but remains marginally behind GPT-4o.

\begin{table}[H]
\centering
\resizebox{0.98\columnwidth}{!}{%
\begin{tabular}{llcccc}
\toprule
\textbf{CPC} & \textbf{Model} & \textbf{O@5} & \textbf{O@10} & \textbf{O@25} & \textbf{Spear} \\
\midrule
\multirow{3}{*}{A61}
    & GPT-4o             & 0.27   & 0.25   & 0.26   & 0.67     \\
    & Llama 3           &  0.26       &   0.25     &   0.26     &    0.63      \\
    & Random  & 0.001 & 0.001 & 0.003 & 0.001   \\
    \midrule
\multirow{3}{*}{G06}
    & GPT-4o             & 0.27   & 0.26   & 0.28   & 0.62     \\
    & Llama 3           & 0.26       &      0.25  &    0.27    &     0.58     \\
    & Random   & 0.001 & 0.002 & 0.004 & -0.0004 \\
    \midrule
\multirow{3}{*}{H04}
    & GPT-4o             & 0.24   & 0.24   & 0.25   & 0.65     \\
    & Llama 3           &  0.23      &   0.22     &   0.25     &   0.61       \\
    & Random   & 0.005  & 0.001 & 0.004 & 0.001   \\
\bottomrule
\end{tabular}
}
\caption{Retrieval similarity between original and LLM generated abstracts across CPC subclasses. Random baseline uses shuffled abstracts. GPT-4o shows the highest retrieval similarity across all subclasses}
\label{tab:retrieval}
\end{table}

\subsection{Qualitative Analysis of Stylistic Features}
\label{sec:qual}
From  standard NLP-based metrics (Table \ref{tab:NLP_Eval}) we observe that most models achieve consistently high scores and the values don't differ much regardless of architecture. These metrics, while useful for surface-level evaluation, appear insensitive to stylistic differences that are critical in the patent domain. To explore further, we conduct a qualitative and linguistic analysis of the generated abstracts. The style metrics as follows. (1) \textbf{Abstract length:} measures verbosity and structural compactness. Longer text may capture more detail invention but risk redundance. On the other hand shorter length may lack specific details. (2) \textbf{Readability:}  shows the linguistic complexity of a text. Higher scores indicatest more intricate sentence structures, often seen in formal or technical writing like patent. (3) \textbf{Passive Voice Usage:}  calculates the use of passive voice in the sentence, which is one of the characteristics of patent language. 

\begin{figure}[ht]
\centering
\includegraphics[width=\columnwidth]{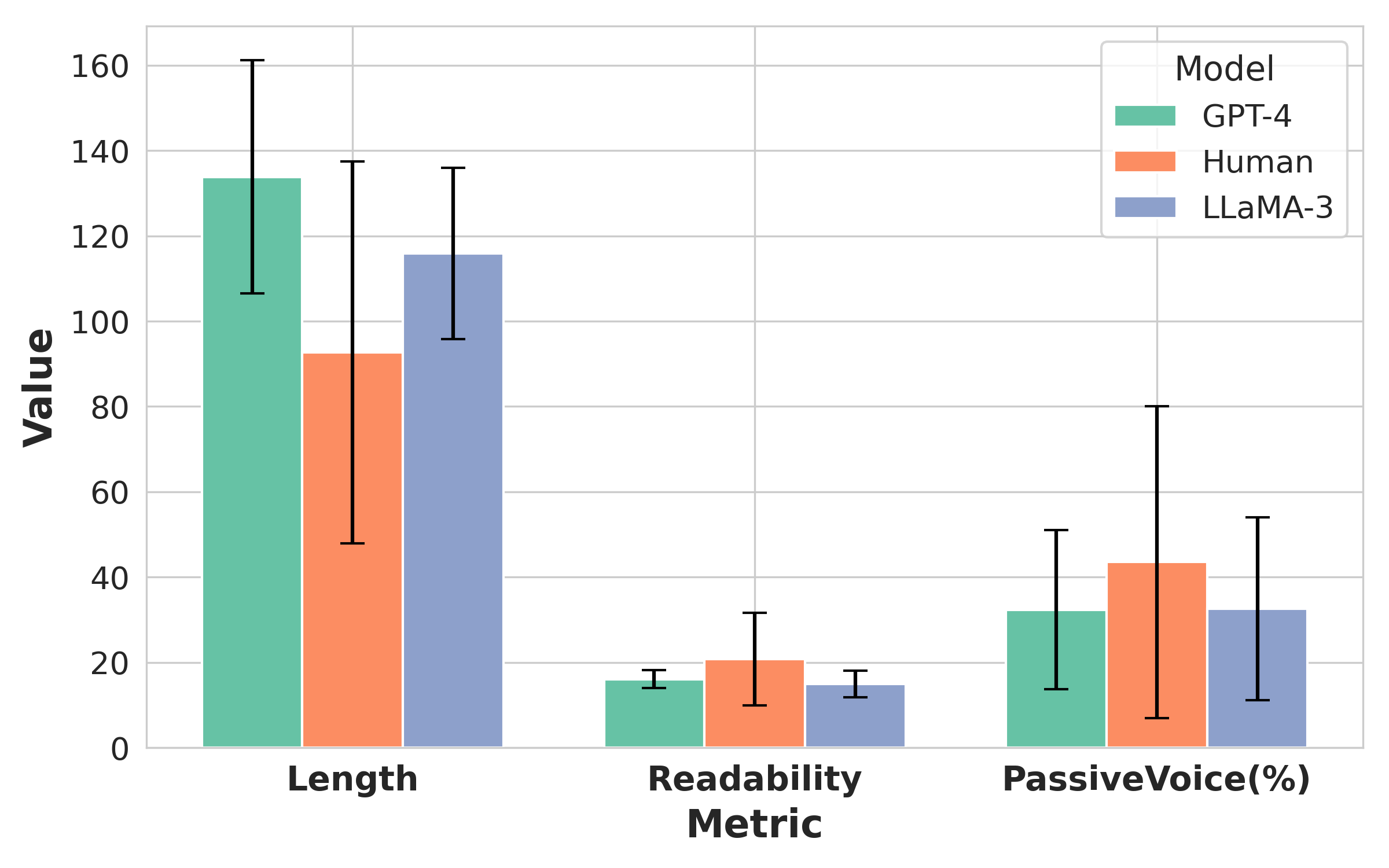}
\caption {\label{fig:qual}Comparison of stylistic metrics (Length, Readability, Passive Voice Usage) across human-written, GPT-4o, and Llama 3 generated abstracts. Human-written abstracts tend to be shorter but exhibit higher readability and greater variation in passive voice usage, while GPT-4o and Llama 3 outputs are more uniform in style, which is expected.
}
\vspace{-3mm}
\end{figure}

We find that, despite similar NLP metric scores, the actual writing styles of the generated abstracts vary meaningfully across models. The outputs generated by GPT-4o are the longest, with an average length of 133.8 ± 27.3, followed by Llama 3 at 115.9 ± 20.1, and human-written abstracts, which are shorter on average (92.7 ± 44.8) but show the most variability. In terms of readability, Human abstracts score the highest (20.8 ± 10.9), while Llama 3 and GPT-4o exhibit lower scores (15.0 ± 3.1 and 16.1 ± 2.1, respectively). Interestingly, GPT-4o and Llama 3 show far less variability in readability than humans. The highest average for passive voice usage is found in human-written abstracts, with an average of 43.6\% ± 36.6\%. This suggests significant variety in tone and grammatical choice. Both Llama 3 and GPT-4o have comparable average usage (32.6\% ± 21.5\% and 32.4\% ± 18.6\%, respectively), but their distributions are more constrained, which indicates a more uniform stylistic template. Figure \ref{fig:qual} shows the barchart. This stylistic analysis demonstrates that traditional NLP metrics alone are insufficient, and supports the need for more domain-aware evaluation in the patent drafting setting.

\section{Discussion}
This work introduces \ourwork, a benchmark designed to evaluate how well LLMs draft patent abstracts. We show that state-of-the-art models like GPT-4o and LLaMA 3 are capable of generating abstracts that are not only accurate, but also comparable to human-written ones in many cases. These models perform consistently well across standard NLP metrics, and they remain reliable even when inputs are noisy. Beyond surface-level similarity, we also find that the generated texts are effective in practical downstream tasks such as patent classification and retrieval. To better understand stylistic tendencies, we include qualitative analyses that highlight how LLMs differ from humans in tone, structure, and writing conventions. Overall, \ourwork offers a tool for studying automated patent drafting. 
Our findings show several practical insights for researchers seeking to use LLMs for patent drafting. Below, we summarize the key takeaways from the benchmark:
\begin{itemize}
\item \textbf{High-quality generation:} State-of-the-art models such as GPT-4o and LLaMA 3 are capable of generating fluent and semantically accurate patent abstracts. Across all three CPC domains, models achieve high BERTScores ($\geq$0.85) that show strong alignment with human-written abstracts.

\item \textbf{Robustness to noisy inputs:} The models, particularly GPT-4o, show stable performance even when the input claim is perturbed with typos, word swaps, or contextual replacements. This highlights their resilience in scenarios where real-world inputs may be imperfect.

\item \textbf{Usefulness in downstream tasks:} The generated abstracts are not only linguistically sound but also functionally useful. In classification and retrieval tasks. LLM-generated abstracts, especially those from GPT-4o, closely match or even outperform original abstracts. This suggests that such outputs can be reliably used in real-world patent analytics pipelines.

\item \textbf{Stylistic  limitations:} Human-written abstracts exhibit greater variability in style, readability, and tone. In contrast, LLM outputs are more uniform and longer on average but less readable. This underscores the need for domain-specific fine-tuning if one wishes to fully replicate expert writing style.
\end{itemize}



\clearpage
\section{Ethical considerations}
The ethical considerations regarding the generation of patents through Large Language Models (LLMs) include the following aspects: 

\begin{itemize}
\item \textbf{Needs for human supervision. }Patent generation should not be fully automated and requires human supervision. Balancing the use of technology with human oversight is important to maintain the quality and integrity of patent applications. Nonetheless, our findings suggest that LLMs could be used as an aid in patent writing.
\item \textbf{Legal issues.} Ethical considerations should also include ensuring that LLM-generated patents comply with legal requirements and regulations of patent laws.
\end{itemize}

\section{Limitations}
This paper addresses a timely subject related to the assistance of AI tools in generating or drafting patents. The dataset and the model used for this study are publicly available. 
While this benchmarking study shows the capability of several open-source LLMs in many different settings of patent abstract generation, it does not go into the details of the implementation of the LLMs that are being deployed in practice. Given the potential impact on the patent system, further exploration of the feasibility and scalability of patent generation might enhance the practical implications of the research.

\clearpage
\bibliographystyle{acl_natbib}
\bibliography{custom}

\clearpage
\appendix

\section{Additional Details on Benchmarking Framewrok}
We show some additional detail of the framework in the following subsections.
\subsection{Prompting Examples}
\label{app:prompting}
In this section, we present the most effective prompt through experimentation with various prompting strategies. For clarity, instructional text is written in black  and additional context or variable information is highlighted in blue.
\begin{tcolorbox}[colback=gray!10, colframe=gray!60!black, 
  title={Zero-shot prompt for abstract generation}]
You are a patent expert. Given the following patent claim, write an informative abstract that captures the key invention, technical purpose, and functionality. 

\textbf{Patent Claim:} \textcolor{blue}{\{Claim\}}.\\
\textbf{Abstract:}
\end{tcolorbox}

\begin{tcolorbox}[colback=gray!10, colframe=gray!60!black, 
  title={Few-shot prompt for abstract generation}]
You are a patent expert. Given the following patent claim, write an informative abstract that captures the key invention, technical purpose, and functionality. \\ 
Example 1: \\
\textbf{Patent Claim:} \textcolor{blue}{\{Claim\}}.\\
\textbf{Patent Abstract:}  
\textcolor{blue}{\{Abstract\}}.\\
Example 2: \\
\textbf{Patent Claim:} \textcolor{blue}{\{Claim\}}.\\
\textbf{Patent Abstract:} 
\textcolor{blue}{\{Abstract\}}.\\
Example 3: \\
\textbf{Patent Claim:}  \textcolor{blue}{\{Claim\}}.\\
\textbf{Patent Abstract:} \textcolor{blue}{\{Abstract\}}.\\
Now, write an abstract for the following claim:
\textcolor{blue}{\{Claim\}}.\\
\textbf{Abstract:}
\end{tcolorbox}

\begin{tcolorbox}[colback=gray!10, colframe=gray!60!black, 
  title={Chain-of-thought prompt for abstract generation}]
You are a patent expert. Given a patent claim, first analyze it step by step to identify the key invention, its technical purpose, and how it functions. Then, based on this reasoning, write a formal abstract.  

Example 1: \\
\textbf{Patent Claim:} \textcolor{blue}{\{Claim\}}.\\
Step-by-step reasoning: \\
\textbf{Patent Abstract:} 
\textcolor{blue}{\{Abstract\}}.\\
Example 2: \\
\textbf{Patent Claim:} \textcolor{blue}{\{Claim\}}.\\
Step-by-step reasoning: \\
\textbf{Patent Abstract:}
\textcolor{blue}{\{Abstract\}}.\\

Now, write an abstract for the following claim:
\textcolor{blue}{\{Claim\}}.\\
\textbf{Abstract:}
\end{tcolorbox}

\subsection{Perturbation Examples }
\label{app:perturbation}
\begin{figure*}[!ht]
  \includegraphics[width=\textwidth]{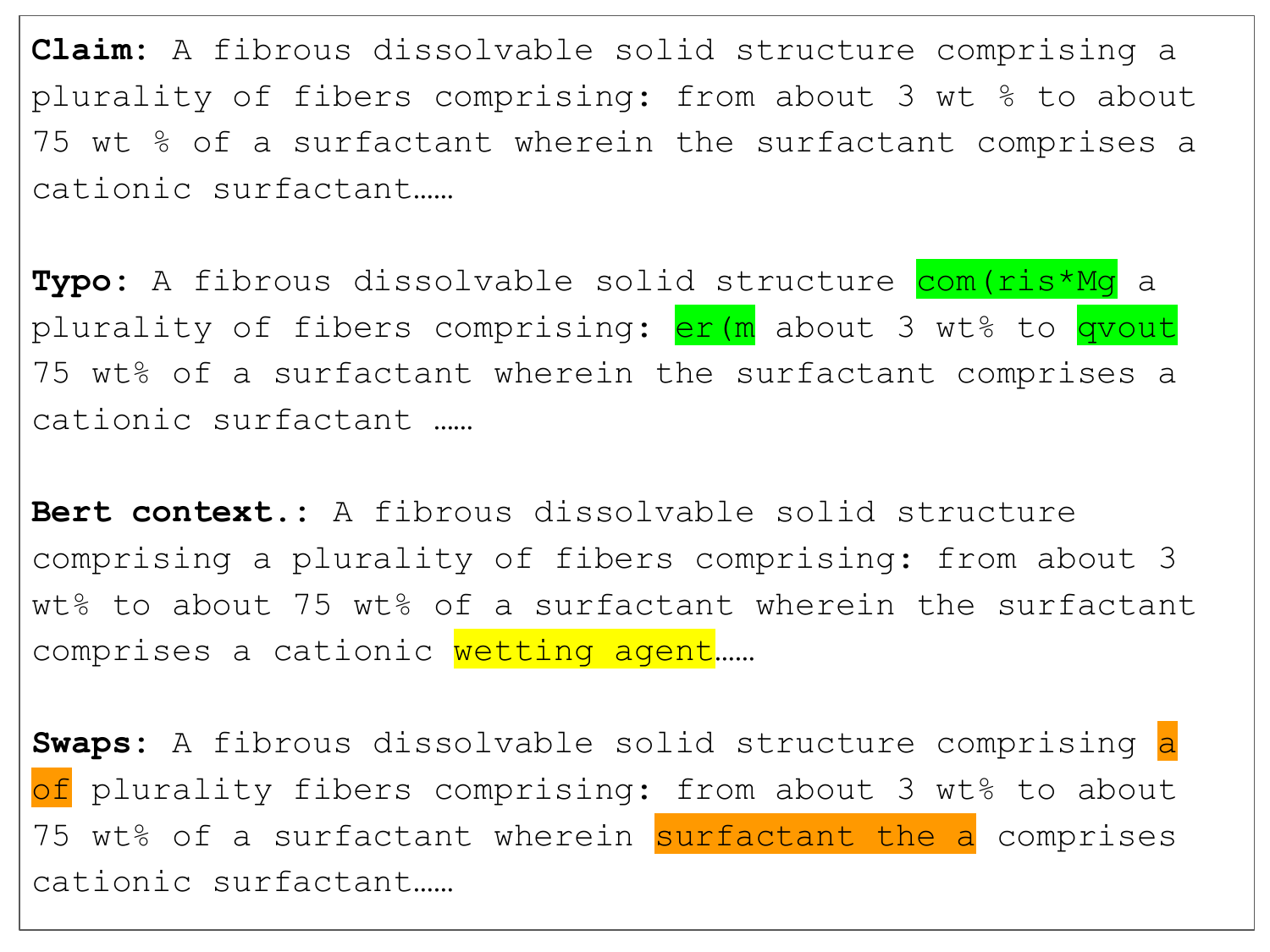}
  \caption{Examples of input perturbations applied to patent claims. The original claim  is modified using three perturbation strategies: (i) Typo, where random characters are injected into words (highlighted in green); (ii) BERT-based context substitution, where a phrase is replaced by a contextually similar term (highlighted in yellow); and (iii) Word swaps, where common word-level reorderings are applied (highlighted in orange). These perturbations simulate noisy or imperfect user inputs to evaluate the robustness of LLM-generated abstracts.}
  \label{fig:pert}
\end{figure*}

One example of three perturbation techniques is shown in Fig. \ref{fig:pert}.

\subsection{Drafting examples}
\begin{figure*}[t]
\centering
\includegraphics[width=\textwidth]{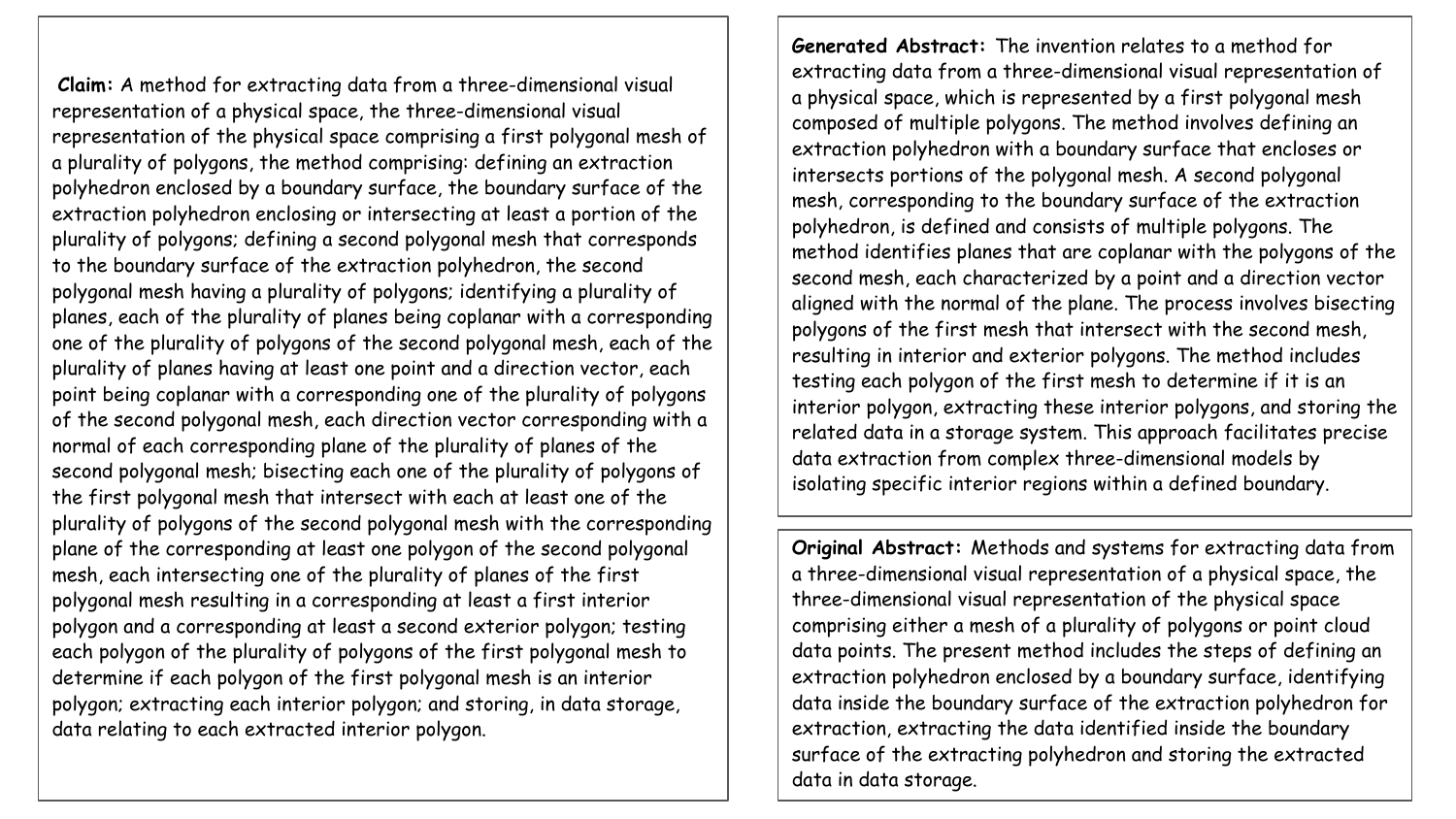}
\caption {\label{fig:example2}Example of  Claim, Generated Abstract (by GPT4o) and Original Abstract.}
\end{figure*}

One example of abstract, claim and generated claim is shown in Fig. \ref{fig:example2}.

\subsection{Experimental setup }
\subsubsection{Patent Classification}
\label{app:claasification_setup}
For our experiments, we employed the all-MiniLM-L12-v2 model from the Sentence-Transformers family as the base encoder for patent classification. We fine-tuned the model with a learning rate of \(2 \times 10^{-5}\), a batch size of 16, and trained for 4 epochs. Evaluation and early stopping were disabled to maintain consistent training duration across datasets. Final results were reported using accuracy, precision, recall, and F1-score. 

\subsubsection{Patent Retrieval}
\label{app:retrieval_setup}
We use all-MiniLM-L6-v2 model to get the embeddings. Abstract embeddings are computed in batches with a maximum sequence length of 256 and a batch size of 32. Cosine similarities are calculated between each query and all other abstracts, excluding self-matches. Top-k rankings are derived from these scores, and retrieval similarity is measured using overlap@k and Spearman rank correlation. All results are averaged over the dataset to ensure robustness.

\section{Data Construction and Preprocessing}
\label{app:data}
We construct our dataset by processing U.S. patent records from the PatentsView, focusing on patents granted in the year 2022. We extract and merge information from two core files: g\_patent and g\_cpc\_current. The g\_patent file provides metadata such as patent ID, title, and abstract and g\_cpc\_current contains the Cooperative Patent Classification (CPC) hierarchy. We primarily focus on class A61 (medical or veterinary science and hygiene), which is among the most frequently granted patent classes. To ensure sufficient representation across technical domains, we retain only those CPC subclasses with at least 1,000 patent instances. This filtering yields a balanced dataset comprising subclasses such as A61B, A61F, A61K, A61L, A61M, A61N, A61P, and A61Q. We also process G06 (computing) and H04 (telecommunications), including G06F, G06K, G06N, G06Q, G06T, G06V, H04B, H04J,H04L, H04M, H04N, H04R, and H04W. A detailed mapping of CPC classes and their descriptions is provided in Table \ref{tab:data_class_subclass_full}. Each entry in the final dataset contains four fields: patent ID, title, abstract, and CPC subclass. After filtering, the dataset consists of approximately 21,000 records, with around 1,000 samples per selected subclass to support balanced training and evaluation for downstream tasks.
\section{Data Analyses}
\label{app:data}
\begin{figure*}[t]
    \centering
    \captionsetup[subfigure]{labelfont={normalsize,bf},textfont={normalsize}}
    \subfloat[original abstract]{\includegraphics[width=0.5\textwidth]{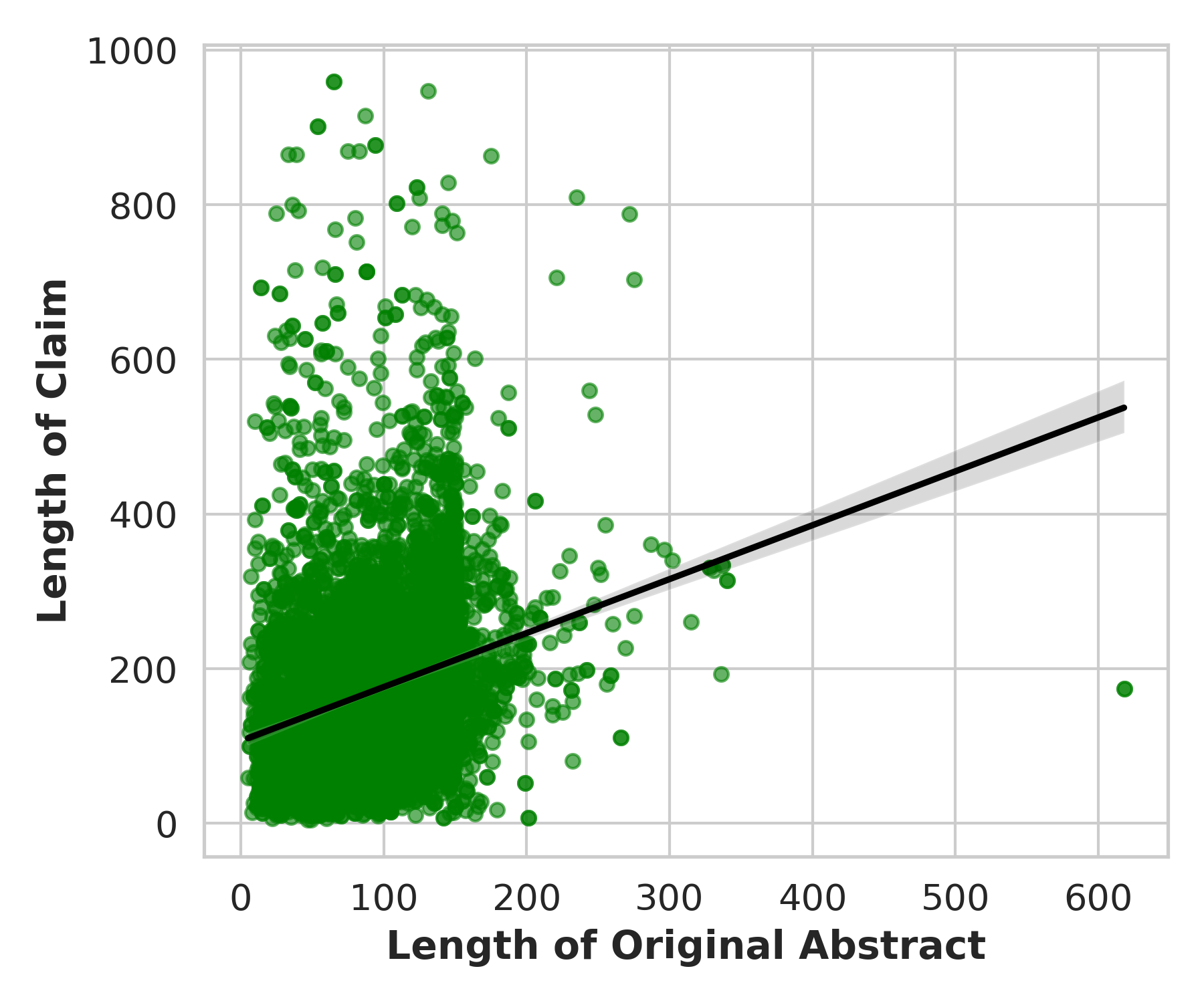}\label{fig:length_org_abs_claim_G06}}
    \subfloat[generated abstract]{\includegraphics[width=0.5\textwidth]{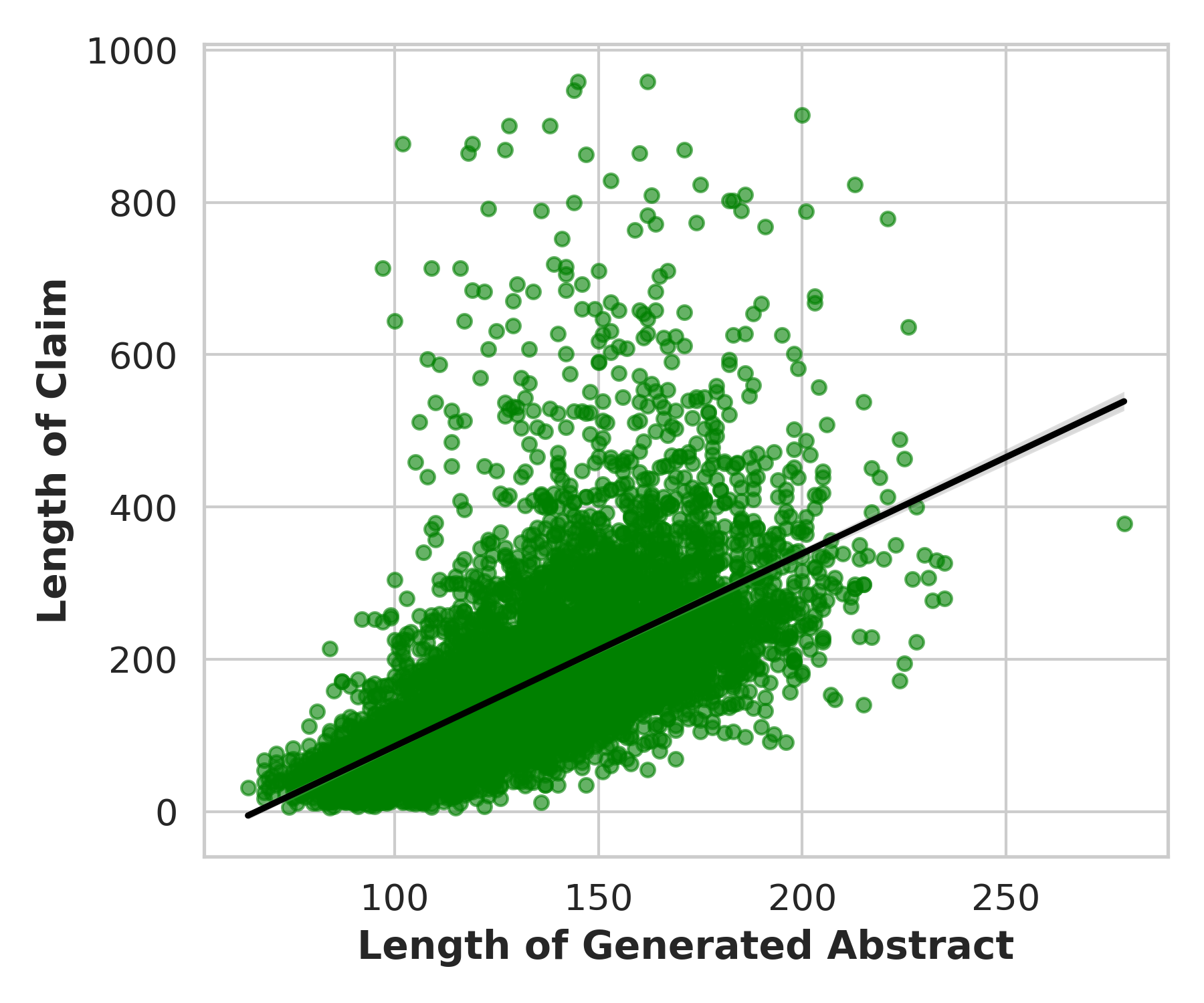}\label{fig:length_gen_abs_claim}}
    \caption{(a) Correlation between the lengths of the original abstract and the first claim where the correlation coefficient is 0.27 (b) Correlation between the lengths of the generated abstract and the first claim where the correlation coefficient is 0.61.
    \label{fig:length_abs_gabs_claim}}
\end{figure*}

To explore how closely abstract length follows the length of the input claim, we analyze the relationship between the number of tokens in the first claim and the corresponding abstract. As shown in Figure~\ref{fig:length_abs_gabs_claim}, human-written abstracts show only a weak correlation with claim length (r= 0.27). This suggests that expert writers don't necessarily adjust abstract length based on how long the claim is. In contrast, the generated abstracts show a much stronger correlation (r= 0.61). This suggests that LLMs are more sensitive to input verbosity, which may contribute to their stylistic consistency.

\begin{figure*}[!ht]
  \includegraphics[width=\textwidth]{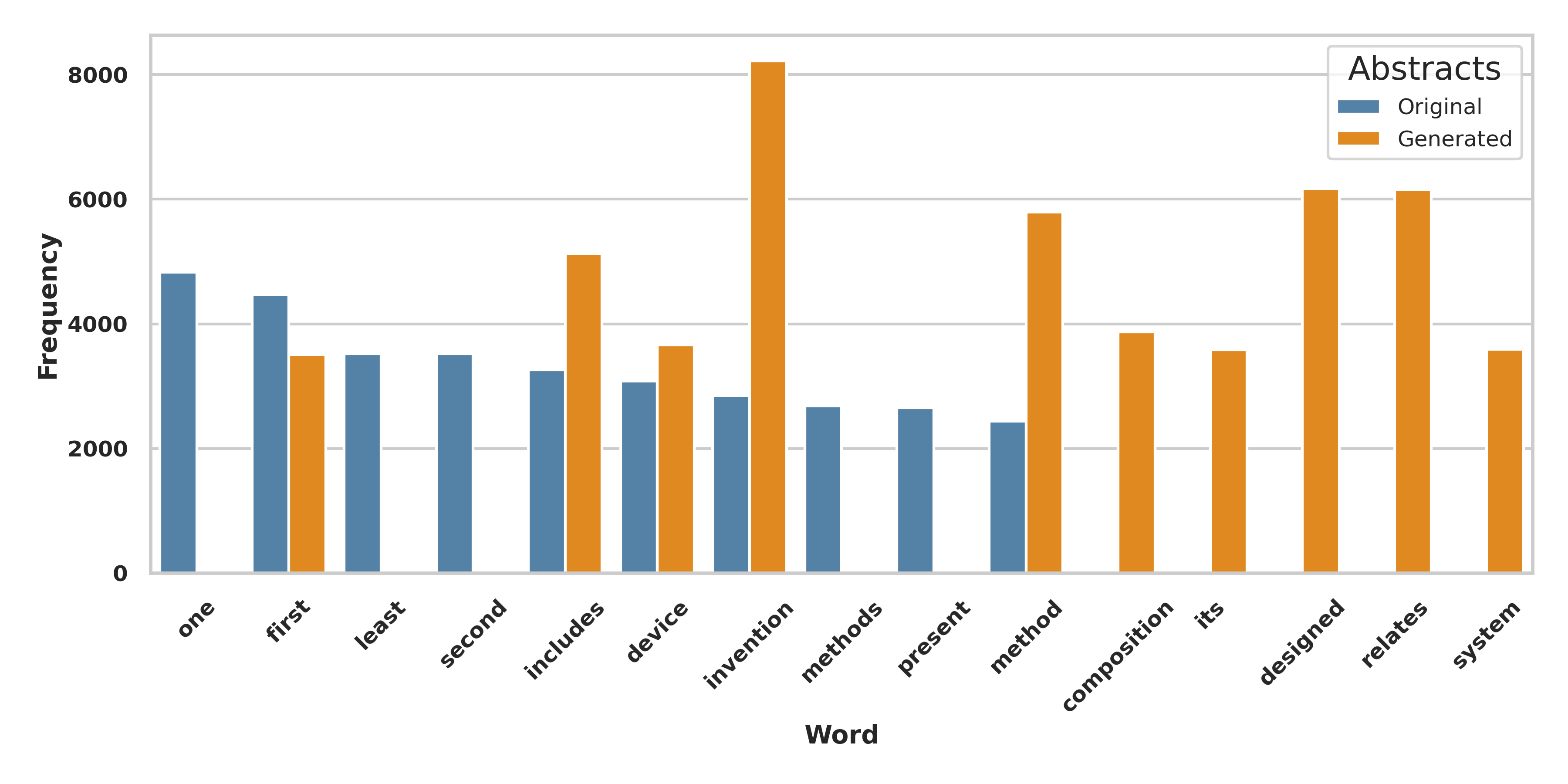}
  \caption{Top 10 most frequent words in original and generated abstracts.}
  \label{fig:topword}
\end{figure*}

We analyze the most frequently used words in both original and generated abstracts. As shown in Figure~\ref{fig:topword}, 5 out of the top 10 words are shared between the two. The word invention appears far more frequently in generated abstracts, along with other structural terms like designed and system.

\section{Inference Time Analyses}
\label{app:time}
We noticed significant differences in how long each language model took to generate the patent abstracts. LLaMA 3 was slower, taking about 39 hours to process the full dataset on an NVIDIA H200 GPU. DeepSeek also took a long time—around 12 hours to generate just 856 samples per subclass—so we only used it on a smaller portion of the data. In contrast, GPT-4o mini and GPT-4.1 were much faster, taking only 21 hours each to complete the entire dataset. These results show the trade-offs between speed, resource use, and model size when choosing a language model for patent drafting at scale.

\section{Patent Classes}
Table~\ref{tab:data_class_subclass_full} provides the CPC subclasses and  descriptions used in our benchmark dataset.
\begin{table*}[ht]
\small
\centering

\resizebox{\textwidth}{!}{%
\begin{tabular}{cccc}
\toprule
\textbf{Classes} & \textbf{Subclasses} & \textbf{Names/Descriptions} \\ \midrule
\textbf{A61}  & A61B & DIAGNOSIS; SURGERY; IDENTIFICATION\\
& A61F &FILTERS IMPLANTABLE INTO BLOOD VESSELS  \\
& A61K & PREPARATIONS FOR MEDICAL, DENTAL OR TOILETRY PURPOSES \\
& A61L & METHODS OR APPARATUS FOR STERILISING MATERIALS OR OBJECTS IN GENERAL  \\
& A61M & DEVICES FOR INTRODUCING MEDIA INTO, OR ONTO, THE BODY \\
& A61N & ELECTROTHERAPY; MAGNETOTHERAPY; RADIATION THERAPY; ULTRASOUND THERAPY \\
& A61P& SPECIFIC THERAPEUTIC ACTIVITY OF CHEMICAL COMPOUNDS OR MEDICINAL PREPARATIONS\\
& A61Q & SPECIFIC USE OF COSMETICS OR SIMILAR TOILETRY PREPARATIONS\\
\hline
\textbf{G06} & G06F &ELECTRIC DIGITAL DATA PROCESSING \\
&G06K & GRAPHICAL DATA READING; PRESENTATION OF DATA; \\ 
&G06N &COMPUTING ARRANGEMENTS BASED ON SPECIFIC COMPUTATIONAL MODELS\\ 
&G06Q &INFORMATION AND COMMUNICATION TECHNOLOGY FOR ADMINISTRATIVE \\ 
&G06T & 	IMAGE DATA PROCESSING OR GENERATION, IN GENERAL\\ 
&G06V & IMAGE OR VIDEO RECOGNITION OR UNDERSTANDING \\ 
\hline
\textbf{H04}  & H04B &TRANSMISSION\\ 
&H04J & MULTIPLEX COMMUNICATION \\
&H04L & TRANSMISSION OF DIGITAL INFORMATION, \\
&H04M & TELEPHONIC COMMUNICATION \\
&H04N & PICTORIAL COMMUNICATION\\
&H04R & LOUDSPEAKERS, MICROPHONES, GRAMOPHONE \\
&H04W & WIRELESS COMMUNICATION NETWORKS  \\

\bottomrule
\end{tabular}%
}
\caption{Table presents the subclasses and their names used in our experiments for patent generation. Note that the class CPC codes have the following names: Classes used for patent generation A61 (Medical or Veterinary Science; Hygiene), G06 (Computing; Calculating or Counting), H04 (Electric Communication Technique). For the sub-classes we use short descriptions as the names are long. The details are available online here: \url{https://www.uspto.gov/web/patents/classification/cpc/html/cpc.html} .}
\label{tab:data_class_subclass_full}
\end{table*}
\label{sec:appendix}

\end{document}